\documentclass[10pt,twocolumn,letterpaper]{article}

\usepackage{cvpr}              

\usepackage{graphicx}
\usepackage{amsmath}
\usepackage{amssymb}
\usepackage{booktabs}
\usepackage{xcolor}

\newcommand{\bg}[2]{%
  \begingroup
  \setlength{\fboxsep}{1pt}
  \raisebox{0pt}[0pt][0pt]{\colorbox{#1}{#2}}%
  \endgroup
}

\definecolor{inputgreen}{RGB}{194, 241, 200}   
\definecolor{flatblue}{RGB}{178, 206, 218}     
\definecolor{lightgreen}{RGB}{124, 202, 132}   
\definecolor{linegray}{RGB}{183, 183, 183}     
\definecolor{renderpurple}{RGB}{201, 139, 195} 

\usepackage[pagebackref,breaklinks,colorlinks]{hyperref}

\usepackage[capitalize]{cleveref}
\crefname{section}{Sec.}{Secs.}
\Crefname{section}{Section}{Sections}
\Crefname{table}{Table}{Tables}
\crefname{table}{Tab.}{Tabs.}

\def\cvprPaperID{*****} 
\def\confName{CVPR}
\def\confYear{2022}

\begin{document}

\title{Workflow-Aware Structured Layer Decomposition for Illustration Production}




\author{
Tianyu Zhang$^{1}$ \quad
Dongchi Li$^{2}$ \quad
Keiichi Sawada$^{2}$\quad
Haoran Xie$^{1,3}$\\
$^{1}$Japan Advanced Institute of Science and Technology\\
$^{2}$Live2D Inc.
$^{3}$Waseda University
}

\maketitle

\begin{abstract}
Recent generative image editing methods adopt layered representations to mitigate the entangled nature of raster images and improve controllability, typically relying on object-based segmentation. However, such strategies may fail to capture the structural and stylized properties of human-created images, such as anime illustrations. To solve this issue, we propose a workflow-aware structured layer decomposition framework tailored to the illustration production of anime artwork. Inspired by the creation pipeline of anime production, our method decomposes the illustration into semantically meaningful production layers, including line art, flat color, shadow, and highlight. To decouple all these layers, we introduce lightweight layer semantic embeddings to provide specific task guidance for each layer. Furthermore, a set of layer-wise losses is incorporated to supervise the training process of individual layers. To overcome the lack of ground-truth layered data, we construct a high-quality illustration dataset that simulated the standard anime production workflow. Experiments demonstrate that the accurate and visually coherent layer decompositions were achieved by using our method. We believe that the resulting layered representation further enables downstream tasks such as recoloring and embedding texture, supporting content creation, and illustration editing. Code is available at:
\url{https://github.com/zty0304/Anime-layer-decomposition}
\end{abstract}

\section{Introduction}
\label{sec:intro}

Making professional anime illustrations follows a meticulous and layered workflow. Designers typically decompose an image into functionally distinct components—line art, flat color, shadow, and highlight—to maintain structural integrity and illumination control. This workflow-aware layering ensures stylistic consistency and enables efficient and controllable editing operations, including localized modifications, recoloring, and relighting. However, despite the rapid advancement of image generation models, state-of-the-art frameworks mainly treat anime imagery as monolithic RGB outputs~\cite{meng2025anidoc,guo2023animatediff, blackforest2024flux,xing2024tooncrafter}. By learning an end-to-end mapping without explicit internal modeling, these models entangle contour geometry, chromatic information, and shading effects within a single feature space. Consequently, while they produce visually impressive results, they may lack the fine-grained controllability required for professional production pipelines and often suffer from inconsistent shading artifacts.

Recent approaches~\cite{lin2026see,yang2025layeranimate, xie2025physanimator}  attempted to bridge this gap through image decomposition and editable representations. However, these layered approaches predominantly focus on object-level segmentation, which separate semantic regions such as foreground and background, or hair, skin, and clothing to facilitate localized editing and style transfer. These methods may be useful for regional editing, but have difficulty reflecting the functional rendering layers used by artists. Within each segment, the line art, base color, and shading remain inseparable, limiting layer-level relighting or large structure-preserving deformation. In addition, physics-based intrinsic decomposition~\cite{careaga2024colorful,careaga2023intrinsic} normally separates images into reflectance and shading—often fails in the anime domain. However, anime illustrations follow stylized, non-photorealistic lighting conventions (e.g., cel-shading) that may not adhere to physical light transport models~\cite{ma2023separating}, leading to semantically meaningless layers and broken stylistic boundaries when processed by traditional approaches.

\begin{figure}[t]
    \centering
    \includegraphics[width=\linewidth]{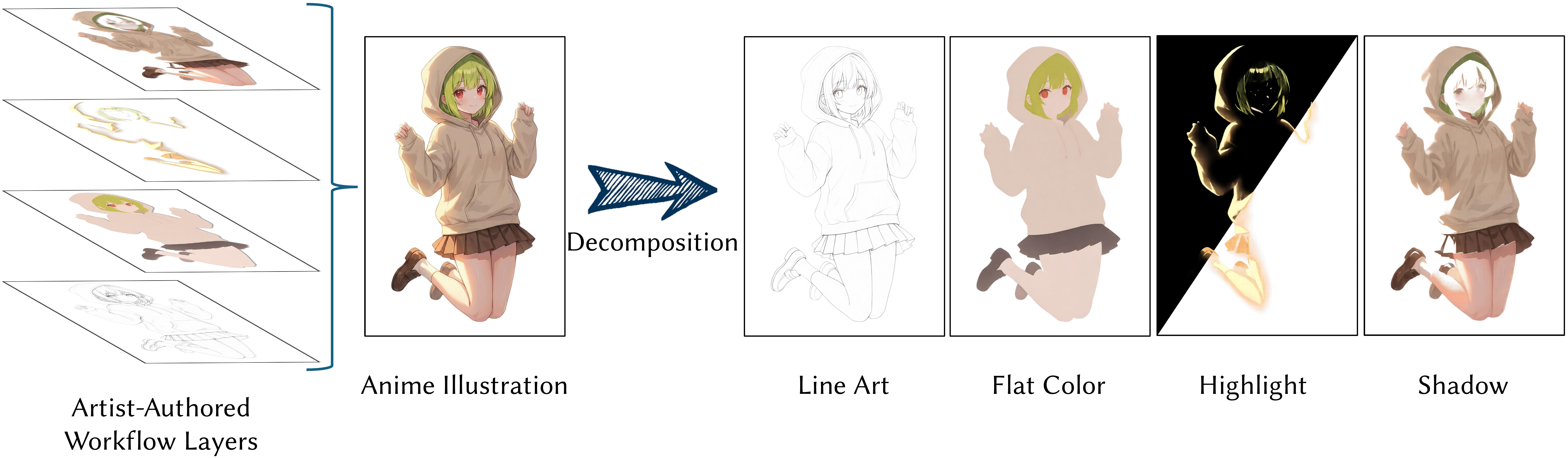}
    \caption{Our framework decomposes anime illustrations into four layers including line art, flat color, highlight, and shadow layers, to align with professional creation workflow.}
    \label{fig:teaser}
\end{figure}

To address these limitations, we propose a workflow-aware structured layer decomposition framework designed specifically for anime illustrations. As shown in Fig.~\ref{fig:teaser}, rather than following object-level or physics-based heuristics, we align our model with professional general drawing practices by decomposing illustrations into four functional  layers: line art, flat color, shadow, and highlight layers. This representation explicitly disentangles structural, chromatic, and illumination components, providing a principled foundation for production-aligned editing and recomposition. We develop a layer-decoupled Diffusion Transformer (DiT) model based on Qwen-Image-Layered~\cite{yin2025qwen} to jointly represent all layers in a shared latent space. 
To mitigate cross-layer interference arising from token mixing, we introduce lightweight layer semantic embeddings that inject explicit layer identity into the token representations, allowing the attention mechanism to distinguish rendering roles during diffusion process. We further adopt parameter-efficient LoRA adaptation to perform layer-specific fine-tuning while freezing the backbone weights, which stabilizes layered learning and prevents over-coupling across components. In addition, we design a layer-aware supervision strategy that applies differentiated objectives to distinct functional layers. The structural layer emphasizes edge sharpness and high-frequency preservation, while illumination layers are regularized to encourage sparsity and inter-layer independence. A compositional consistency constraint is further imposed to ensure that the aggregated decomposition remains globally faithful. Together, this representation-level, parameter-level, and objective-level decoupling enables stable, controllable, and production-aligned four-layer decomposition.

We evaluate the proposed method through both qualitative comparisons and quantitative metrics. The robustness and generalization capability of our method are further verified for production-level illustration decomposition scenarios. We also apply our method to downstream tasks to verify the practical utility of our method. Our main contributions are listed as follows:

\begin{itemize}
    \item A workflow-aware representation that formalizes anime decomposition into four functional rendering layers (line art, flat color, highlight, shadow).
    \item A layer-decoupled DiT model with layer semantic embeddings and layer-wise supervision for controllable layer decomposition.
    \item A dataset of high-quality anime illustrations with aligned functional layers to support future research in structured image decomposition.
\end{itemize}

\section{Related Work}
\label{sec:rw}

\subsection{Anime Synthesis}
With the rapid advancement of artificial intelligence, neural networks have been increasingly applied across various stages of illustration and anime production, such as colorization~\cite{liu2025manganinja, meng2025anidoc}, editing ~\cite{zhang2023adding,blackforest2024flux, batifol2025flux}, deformation~\cite{xie2025physanimator}, inbetweening~\cite{brodt2024skeleton,li2021deep}, and animation generation~\cite{yang2025layeranimate, guo2023animatediff, xing2024tooncrafter}. Despite these progresses, existing methods still fall short of meeting the rigorous and fine-grained standards required in industrial production pipelines.

To achieve high-quality and controllable results, recent research has focused extensively on disentangling visual elements and incorporating explicit guidance signals. A primary strategy is to extract line art or sketches as dense structural guidance for generative models, thereby preserving geometric consistency during synthesis~\cite{xing2024tooncrafter}. To further enhance controllability, several approaches segment characters into distinct semantic parts and inpaint occluded or hidden anatomy to construct manipulable proxies~\cite{yang2025layeranimate,lin2026see}. Such structural disentanglement enables the application of skeletonization and deformation algorithms to animate static illustrations in a smooth and coherent manner~\cite{ou2024body,qiao2024cartoonnet}. Additionally, contemporary frameworks leverage physics-based priors to emulate fluid dynamics effects in animation, such as those governing smoke and clothing motion~\cite{xie2025physanimator,chang2025diffsmoke}.

Despite improved controllability, disentanglement-based methods exhibit part misalignment and color inconsistency~\cite{meng2025anidoc,liu2025manganinja} under large pose discrepancies between two frames. This stems from their neglect of the artist-defined stratified structure, which comprises line art, flat colors, shadows, and highlights.

\subsection{Layer Decomposition}

Effective image disentanglement is essential for enabling controllable editing and high-quality generation. By decomposing a raster image into structured components, users can achieve inherent editability while avoiding semantic drift.

Early approaches extract color palettes using geometric structures such as convex hulls or polyhedrons~\cite{tan2015decomposing, wang2019improved}, which are later extended by neural-network-based palette extraction~\cite{akimoto2020fast} and soft color segmentation techniques~\cite{aksoy2017unmixing}. To support digital painting workflows, Koyama et al.~\cite{koyama2018decomposing} further generalize palette decomposition to handle non-linear blending modes such as multiply and screen. Recent work focuses on semantic-level decomposition, aiming to separate images into meaningful structural components. For natural images, prior methods disentangle foreground/background layers, recover occlusion ordering~\cite{zhan2020self}, or separate visual effects such as shadows~\cite{yang2025generative}. Diffusion-based approaches further enable layer-wise generation and editing~\cite{huang2025dreamlayer, pu2025art, huang2024layerdiff}. In graphic design scenarios, systems such as LayerD, OmniPSD, and CLD~\cite{suzuki2025layerd, liu2025omnipsd, liu2025controllable} decompose raster graphics into semantic layers including typography, geometric elements, and images, while recent generative models like Qwen-Image-Layered~\cite{yin2025qwen} directly support multi-layer generation. 

Despite these advances, semantic decomposition methods remain limited when large structural edits or significant camera depth changes are required, as shading, lighting, and line-art are tightly entangled with object semantics. Meanwhile, vectorization-based approaches either fail to handle complex shading and illumination~\cite{wu2025layerpeeler} or produce overly dense vector representations that are difficult to integrate into practical production pipelines~\cite{xing2025svgdreamer++}.

\begin{figure*}[t]
    \centering
    \includegraphics[width=\linewidth]{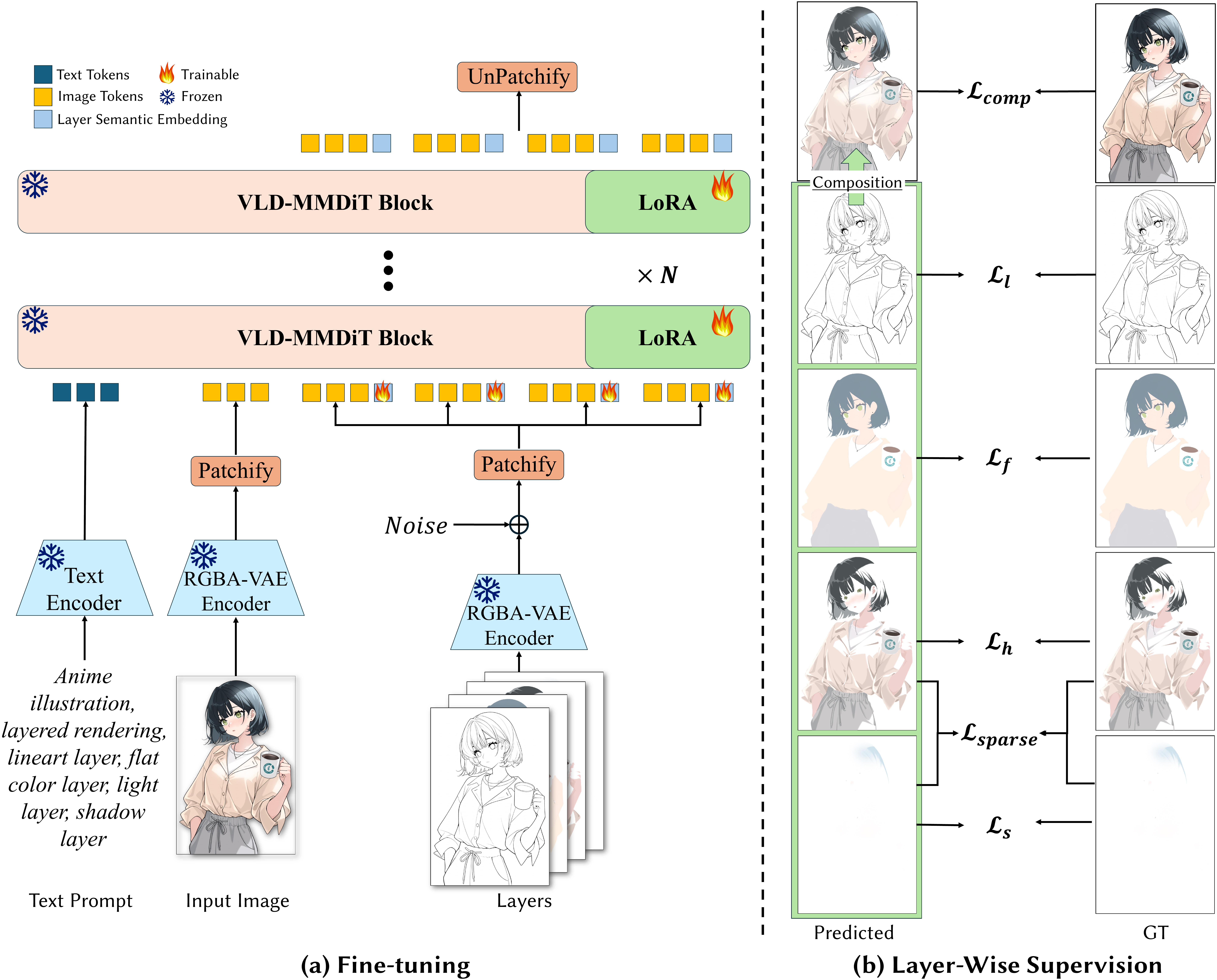}
    \caption{Overview of our anime illustration decomposition framework. (a) We enhance the Qwen-Image-Layered~\cite{yin2025qwen} with layer semantic embeddings and a multi-faceted Layer-wise supervision loss, to improve disentanglement and rendering quality.}
    \label{fig:frame}
\end{figure*}

\section{Preliminaries}
\label{sec:prelim}

\textbf{RGBA-VAE.}
Qwen-Image-Layered~\cite{yin2025qwen} introduces an RGBA-VAE that jointly encodes RGB and RGBA images into a shared latent space, narrowing the distribution gap between RGB inputs and RGBA outputs. 
It extends the first convolution layer of the encoder and the last convolution layer of the decoder from three to four channels. 
To preserve RGB reconstruction ability during initialization, the alpha-channel parameters are initialized as
\begin{equation}
W^{0}_{E}[:,3,:,:,:]=0, \qquad 
W^{l}_{D}[3,:,:,:,:]=0, \qquad 
b^{l}_{D}[3]=1 .
\end{equation}
After training, both RGB images and RGBA layers are encoded into the same latent space, and each RGBA layer is encoded independently without compression along the layer dimension.

\noindent
\textbf{VLD-MMDiT.}
Given RGBA layers $L$, their latent representation is $x_0 = \mathcal{E}(L)$. 
Following Rectified Flow, noise $x_1$ and timestep $t$ are sampled and the intermediate state and velocity are defined as
\begin{equation}
x_t = t x_0 + (1-t)x_1, \qquad
v_t = \frac{d x_t}{dt} = x_0 - x_1 .
\end{equation}
where $x_1 \sim \mathcal{N}(0,I)$ and $t \sim \mathcal{U}(0,1)$. The model predicts the velocity $v_\theta(x_t,t,z_I,h)$ conditioned on the encoded input image $z_I$ and text embedding $h$, and is trained using
\begin{equation}
\mathcal{L} =
\mathbb{E}_{(x_0,x_1,t,z_I,h)\sim\mathcal{D}}
\|v_\theta(x_t,t,z_I,h)-v_t\|_2^2 .
\end{equation}
where $\mathcal{D}$ denotes the training data distribution.

VLD-MMDiT applies multimodal attention over text tokens and visual tokens, and introduces Layer3D RoPE with a layer index to support decomposition with a variable number of layers.

\section{Workflow-Aware Layer Decomposition}
\label{sec:method}

Although Qwen-Image-Layered~\cite{yin2025qwen} demonstrates effective image decomposition, its object-oriented decomposition is inadequate for the functional layers required in the anime illustration workflow. Since layers such as line art and lighting possess unique properties, we propose layer semantic embeddings to mitigate cross-layer interference along with a layer-wise supervision to maintain both structural fidelity and global consistency.

\subsection{Layer Semantic Embedding}
\label{sec:lse}

As shown in Fig.~\ref{fig:frame} (a), four rendering layers (lineart, flat color, highlight, and shadow) are sequentially packed into a latent token sequence. However, the transformer backbone does not explicitly distinguish which tokens belong to which layer. This implicit representation often leads to cross-layer interference and unstable gradient coupling during training, degrading layer disentanglement.

To address this issue, we introduce a lightweight layer semantic embedding (LSE) that injects explicit layer identity information into the latent tokens. Let $x_t \in \mathbb{R}^{B \times T \times C}$ denote the packed latent representation at diffusion time $t$, where $B$ is the batch size, $T$ is the token length, and $C$ is the hidden dimension of the transformer. Since the four layers are sequentially packed along the token dimension, we uniformly partition the sequence into four contiguous segments of equal length $T/4$.

We define a learnable embedding matrix $E \in \mathbb{R}^{4 \times C}$
where each row $e_i \in \mathbb{R}^{C}$ corresponds to the semantic embedding of layer $i$. The embeddings are initialized from $\mathcal{N}(0, 0.02^2)$ and jointly optimized with LoRA parameters during training. For the $i$-th token segment $\mathcal{I}_i$, we inject the layer embedding via additive bias:
\begin{equation}
\tilde{x}_t[:, \mathcal{I}_i, :] = x_t[:, \mathcal{I}_i, :] + e_i
\end{equation}
where index $i \in \{1,2,3,4\}$ that 1, 2, 3, and 4 correspond to line art, flat color, light, and shadow, respectively. This mechanism is analogous to segment embeddings in language models, providing explicit layer-type signals without modifying the transformer architecture or attention topology. The module introduces only $4C$ additional parameters, which is negligible compared to the size of the backbone. During training, the base transformer weights remain frozen, and only LoRA adapters and layer embeddings are updated.

\subsection{LoRA Fine-tuning}

We fine-tune a pretrained Qwen-Image-Layered~\cite{yin2025qwen} transformer using parameter-efficient LoRA adaptation. Let $W_0$ denote the original attention projection weight. LoRA decomposes the weight update into a low-rank residual:
\begin{equation}
W = W_0 + AB
\end{equation}
where $A \in \mathbb{R}^{d \times r}$, $B \in \mathbb{R}^{r \times k}$, $W_0 \in \mathbb{R}^{d \times k}$, and $r$ is the rank. In our implementation, we set $r=32$ and insert LoRA modules into the attention projections. All backbone weights $W_0$ are frozen, and only LoRA parameters and the layer semantic embeddings are optimized.

\subsection{Layer-Wise Supervision}

Under layered decomposition, a single global loss is insufficient for the following reasons: (1) different rendering layers exhibit distinct statistical properties; (2) lineart requires structural sharpness, which is often blurred under MSE supervision; and (3) illumination components are inherently sparse and prone to leakage across layers.
Therefore, we supervise the model in a set of layer-aware losses. Both prediction and target are partitioned into 4-layer segments:
\begin{equation}
\hat{v} = [\hat{v}^{(1)}, \hat{v}^{(2)}, \hat{v}^{(3)}, \hat{v}^{(4)}], \quad
v = [v^{(1)}, v^{(2)}, v^{(3)}, v^{(4)}].
\end{equation}
Let $\hat v = v_\theta(x_t,t,z_I,h)$ denote the predicted velocity. As shown in Fig.~\ref{fig:frame} (b), for the lineart layer, we apply an $L_1$ loss to preserve sharp structural details better:
\begin{equation}
\mathcal{L}_l = \|\hat{v}^{(1)} - v^{(1)}\|_1
\end{equation}
For flat color ($\mathcal{L}_{\text{f}}$), highlight ($\mathcal{L}_{\text{h}}$), and shadow ($\mathcal{L}_{\text{s}}$) layers, we adopt mean squared error (MSE loss):
\begin{equation}
\mathcal{L}_{\{f,h,s\} } = 
\frac{1}{N} \|\hat{v}^{(j)} - v^{(j)}\|_2^2  \quad j \in \{2,3,4\}
\end{equation}
where $N$ denotes the number of elements in each layer segment. To encourage sparse and disentangled illumination decomposition, we introduce an $L_1$ regularization term on light and shadow predictions:
\begin{equation}
\mathcal{L}_{\text{sparse}} = \|\hat{v}^{(3)}\|_1 + \|\hat{v}^{(4)}\|_1
\end{equation}
Furthermore, we enforce compositional consistency to prevent layer-wise drift:
\begin{equation}
\mathcal{L}_{\text{comp}} =
\frac{1}{N}
\left\|
\sum_{i=1}^{4} \hat{v}^{(i)} -
\sum_{i=1}^{4} v^{(i)}
\right\|_2^2 
\end{equation}
This constraint ensures that the aggregated residual remains faithful to the ground truth, avoiding degenerate solutions where individual layers are locally correct but globally inconsistent.

The final training objective is
\begin{equation}
\mathcal{L} =
\lambda_l \mathcal{L}_l +
\lambda_f \mathcal{L}_f +
\lambda_h \mathcal{L}_h +
\lambda_s \mathcal{L}_s +
\lambda_{sparse} \mathcal{L}_{\text{sparse}} +
\lambda_{comp} \mathcal{L}_{\text{comp}}
\end{equation}
where $\{\lambda\}$ are scalar weights. We assign higher weights to the structural line art term and compositional consistency, while applying moderate sparsity regularization to balance decomposition fidelity and editability.

\section{Experiments}

\subsection{Dataset Construction}

\begin{figure}[t]
    \centering
    \includegraphics[width=\linewidth]{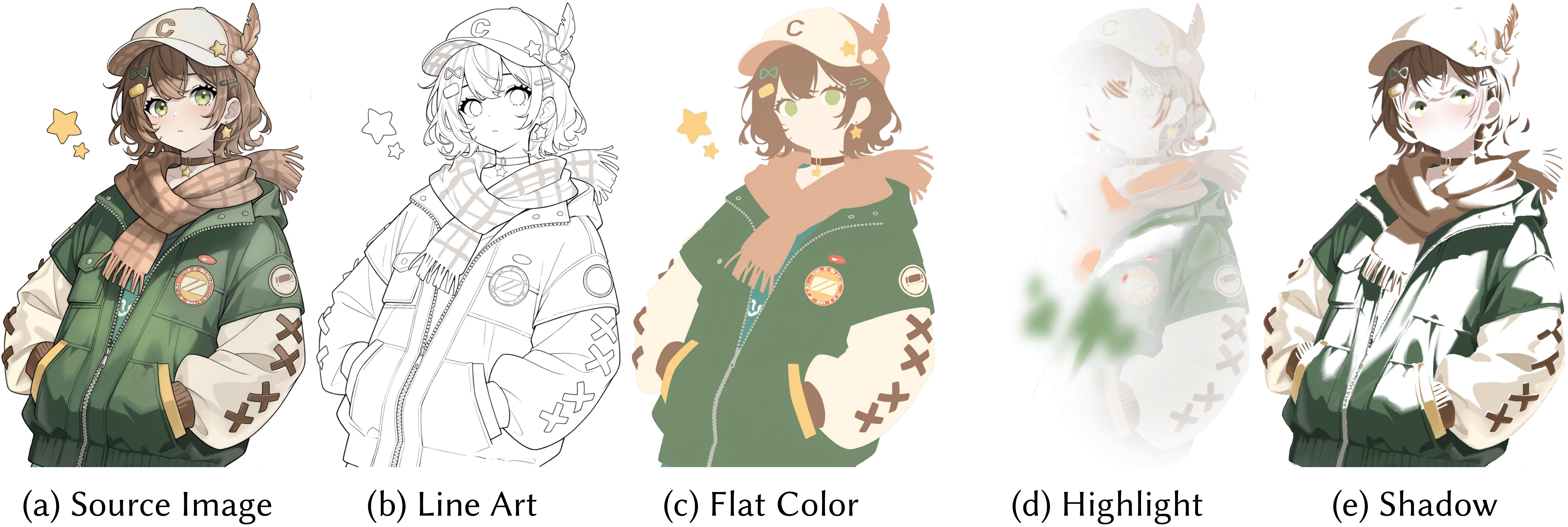}
    \caption{Overview of our anime illustration decomposition dataset. Each sample consists of (a) a source image and its manually decomposed layers: (b) line art, (c) flat color, (d) highlight, and (e) shadow. 
    }
    \label{fig:dataset}
\end{figure}

\begin{figure*}[t]
    \centering
    \includegraphics[width=\linewidth]{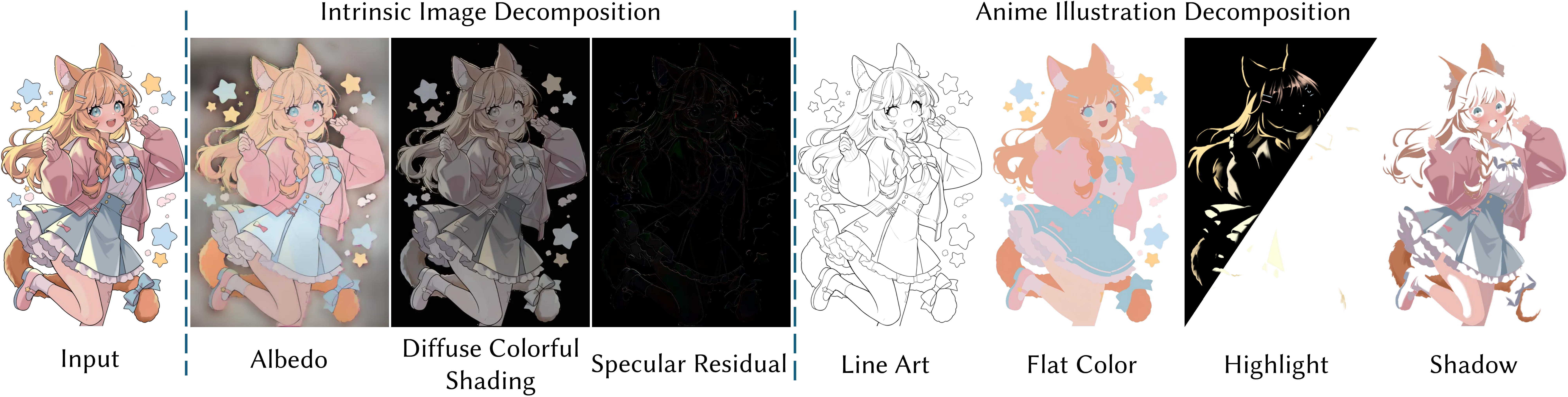}
    \caption{Comparison between intrinsic image decomposition and our anime illustration decomposition. For the same input, physics-based methods result in entangled representations of line and color. Our method successfully extracts the line art, flat color, highlight, and shadow layers consistent with artistic workflows.}
    \label{fig:real}
\end{figure*}

Obtaining high-quality, layer-decomposed anime illustrations is highly challenging, as such assets are time-consuming to produce and existing professional resources are often restricted by copyright, making them difficult to use for research purposes. 
To address this limitation, we construct a four-layer anime illustration dataset to support model training. 
As shown in Fig.~\ref{fig:dataset}, each sample in our dataset is organized as a five-tuple consisting of: 
(a) source image representing the final character appearance, along with its corresponding decomposed layers, including 
(b) line art layer, 
(c) flat color layer, 
(d) highlight layer, and 
(e) shadow layer.

We first generate single-character anime images using Stable Diffusion~\cite{rombach2022high} and remove their backgrounds. 
However, due to inconsistent lighting effects and color artifacts, these generated results are not directly suitable as high-quality source images. 
To ensure aesthetic consistency and professional quality, we collaborated with two experienced character creators: one serves as the primary artist and the other as a supervisor. 
Through iterative refinement and mutual feedback, they carefully adjusted each image to produce a visually coherent and high-quality final source image.

For each finalized source image, the artist reconstructs the complete illustration in a layered PSD file, explicitly decomposing it into the four constituent layers (line art, flat color, highlight, and shadow). 
These layers are required to recompose the source image when combined accurately. 
Throughout this process, the supervisor reviews each stage and provides detailed feedback to ensure precision and visual fidelity.
In total, our dataset contains $45$ carefully curated and professionally refined five-tuple samples with high aesthetic quality and precise layer alignment for model training.
For validation, as obtaining a large volume of copyright-cleared and meticulously layered images remains challenging, we employ a testing set consisting of 108 generated images, each rigorously inspected by professional artists to ensure quality and structural integrity.
For both training and testing, all images are exported in RGBA format and resized to a unified resolution of $768 \times 1152$.

\begin{figure*}[!t]
    \centering
    \includegraphics[width=\linewidth]{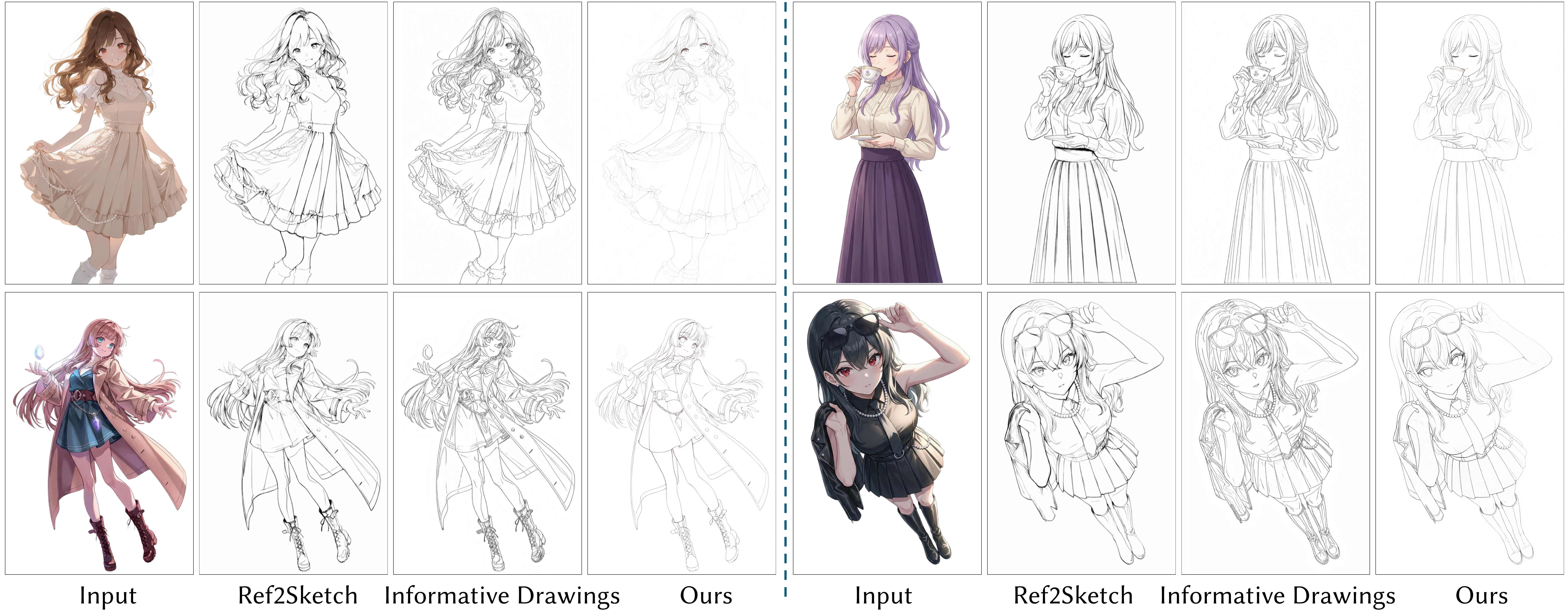}
    \caption{Qualitative comparison of line art results between our method and informative drawings~\cite{chan2022learning} and Ref2Sketch~\cite{seo2023semi}. }
    \label{fig:line}
\end{figure*}

\begin{figure*}[!t]
    \centering
    \includegraphics[width=\linewidth]{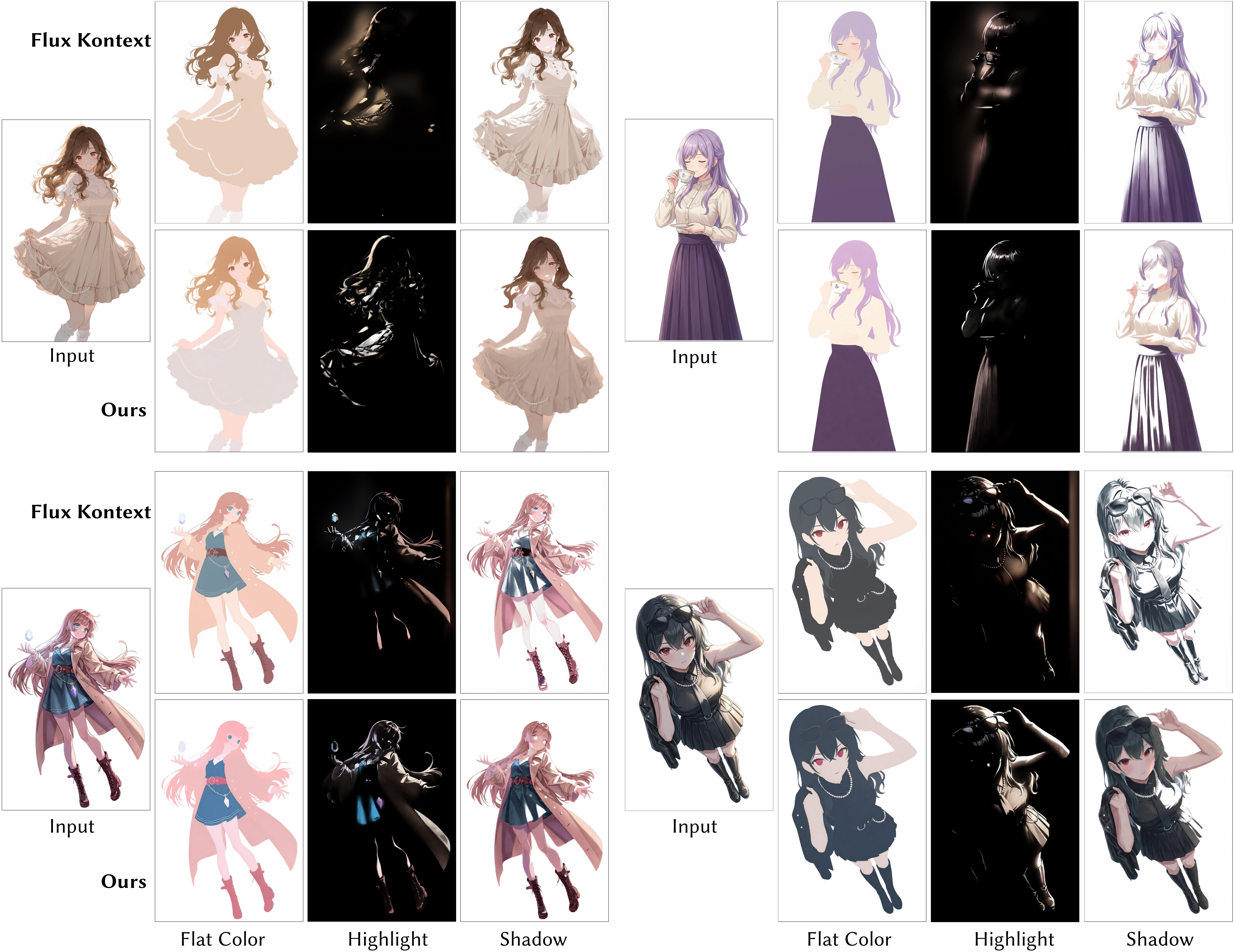}
    \caption{Qualitative comparison of flat color, highlight, and shadow results between our method and Flux Kontext~\cite{batifol2025flux}. }
    \label{fig:others}
\end{figure*}

\subsection{Implementation Details}

We trained our method based on Qwen-Image-Layered\cite{yin2025qwen}. LoRA fine-tuning was adopted with a learning rate of $1 \times 10^{-4}$, rank $r = 32$, and bfloat16 precision. A fixed text prompt was used during both training and inference to ensure consistent conditioning:
“\textit{Anime illustration, layered rendering, lineart layer, flat color layer, light layer, shadow layer}”.
The overall training process takes approximately 40 hours on a single RTX PRO 6000 Blackwell GPU. At inference time, generating a single image requires about 2 minutes on the same GPU.
We additionally applied layer-wise supervision with weights set to $\lambda_{l}=2.5$, $\lambda_{sparse}=0.08$, and $\lambda_{comp}=1.8$, while $\lambda_{f}, \lambda_{h}, \text{ and } \lambda_{s}$ are all set to $1.0$.

\subsection{Comparison with Existing Methods}

\begin{figure*}[t]
    \centering
    \includegraphics[width=\linewidth]{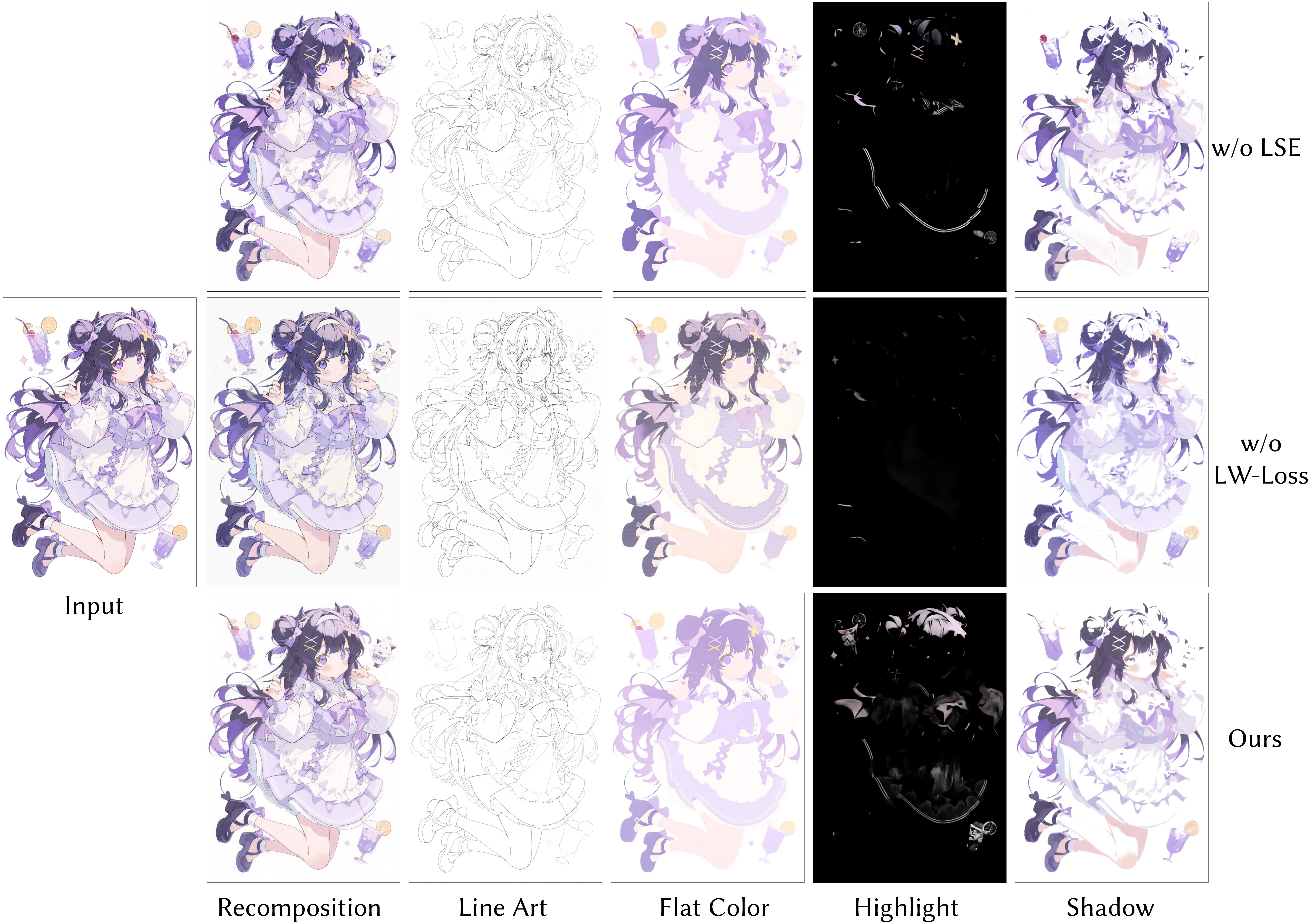}
    \caption{Qualitative comparison of ablation study. We present the decomposition and recomposition results of different variants.}
    \label{fig:ablation}
\end{figure*}

Given the novelty of our task, there is no prior method that enables a complete direct comparison. 
Numerous intrinsic image decomposition methods have been proposed for natural images based on physical image formation models, typically decomposing an image into components such as albedo and shading~\cite{careaga2024colorful}. While the albedo component shares some semantic overlap with the flat color layer in anime, these physically-grounded formulations do not directly align with the stylized layering conventions inherent to anime production (as shown in Fig.~\ref{fig:real}).

Therefore, to contextualize our approach within the state of the art methods, we select representative baselines per layer and conduct layer-wise evaluations. Specifically, we adopt Informative Drawing~\cite{chan2022learning} and Ref2Sketch~\cite{seo2023semi} as baselines for the line art layer. For the flat color, highlight, and shadow layers, we adopt FLUX Kontext~\cite{batifol2025flux}, a recent in-context editing model designed for multi-input conditioning through image concatenation. We further fine-tune the Flux Kontext with LoRA on our dataset to ensure a fair comparison.

\textbf{Qualitative Results.} In Fig. \ref{fig:line}, we compare the line arts generated by Ref2Sketch, Informative Drawings, and our method. Both Ref2Sketch and Informative Drawings often generate line arts containing many extraneous lines, such as shading boundaries and tonal transition strokes. These lines do not belong to the character’s structural contours and introduce visual clutter, making the results difficult to be directly utilized in typical anime workflows. In contrast, our method generates clean line art that primarily preserves the structural contours of the character. Because our framework explicitly decouples line art from shading layers, non-structural strokes are effectively removed.

Additionally, as shown in Fig~\ref{fig:others}, the fine-tuned Flux Kontext can generate the corresponding layers. However, the results often exhibit noticeable color shifts, as the model still struggles to fully account for the fact that the input colors are affected by shading and illumination. In addition, the generated highlights frequently present diffused and feathered boundaries. Moreover, Flux Kontext only supports one-to-one image generation, which prevents cross-supervision among layers, such as the introduction of composition losses. In contrast, our method accurately captures light, shadow, and flat color layers while maintaining high consistency with the input image.

\textbf{Quantitative Results.} Due to the lack of ground truth (GT) layer decompositions, direct quantitative evaluation is not feasible. We evaluate reconstruction quality by recomposing the predicted layers and comparing the result with the input image. We report Peak Signal-to-Noise Ratio (PSNR)~\cite{wang2004image} for image fidelity, Structural Similarity Index Measure (SSIM)~\cite{wang2004image} for structural similarity, and Local Mean Squared Error (LMSE)~\cite{grosse2009ground} for local structural errors. As shown in Table~\ref{tab:com}, our method consistently achieves the best performance across all metrics. This demonstrates that our method can effectively decompose the illustration into four distinct layers while maintaining high fidelity to the input.

\begin{table}
        \centering
        \caption{Quantitative evaluation results. Recomposition variants use lineart from: (a) Informative Drawings~\cite{chan2022learning}; (b) Ref2Sketch~\cite{seo2023semi}. In both cases, the flat color, shadow, and light layers are sourced from FLUX kontext~\cite{batifol2025flux}.}
        \label{tab:com}
        \begin{tabular}{c c c c}
        \toprule 
        & PSNR($\uparrow$) & SSIM($\uparrow$) & LMSE($\downarrow$)\\ \hline
        (a) & 11.37 & 0.6725  & 0.0794\\
        (b) & 11.45 & 0.6974  & 0.0788\\
        Ours & \textbf{27.79} & \textbf{0.9254} & \textbf{0.0017}\\
        \bottomrule 
        \end{tabular}
\end{table}


\subsection{Ablation Study}

We designed two variants for the ablation study to analyze the individual contributions of each core component:

\begin{itemize}
    \item w/o LSE: We eliminated the Layer-Semantic Embedding (LSE) module from our full framework while keeping the loss for each layer unchanged. This variant is designed to evaluate the impact of LSE on semantic guidance between layers.
    \item w/o LW-Loss: We replaced all specialized loss functions with the canonical MSE loss used in the base model. This variant serves to investigate the efficacy of layer-wise supervision in facilitating structural and color decoupling.
\end{itemize}

As shown in Fig. \ref{fig:ablation}, removing the LSE leads to interference between layers since all layers are inferred jointly without explicit disentanglement. A notable example can be observed in the first row: the predicted highlight layer contains visible structures from the line art and flat color layers, indicating entanglement between representations. When the layer-wise supervision is removed, the line art and flat color layers exhibit noticeable color deviations, and the highlight layer, due to its sparse nature, becomes difficult for the model to capture reliably. Quantitatively, as reported in the Table~\ref{tab:abla}, our method achieves the highest PSNR and SSIM and the lowest LMSE, demonstrating superior visual fidelity and better preservation of structural details.

\begin{table}
    \centering
    \caption{Quantitative comparison of the ablation study. ``w/o LSE'' represents the full method without layer semantic embeddings and ``w/o LW-Loss'' represents the method without layer-wise supervision}
    \label{tab:abla}
    \begin{tabular}{c c c c}
    \toprule 
    & PSNR($\uparrow$) & SSIM($\uparrow$) & LMSE($\downarrow$)\\ \hline
    w/o LSE & 27.38 & 0.9198  & 0.0019\\
    w/o LW-Loss & 24.75 & 0.8845  & 0.0038\\
    Ours & \textbf{27.79} & \textbf{0.9254} & \textbf{0.0017}\\
    \bottomrule 
    \end{tabular}
\end{table}

\subsection{Generalization with Downstream Tasks}

\begin{figure*}[!t]
    \centering
    \includegraphics[width=\linewidth]{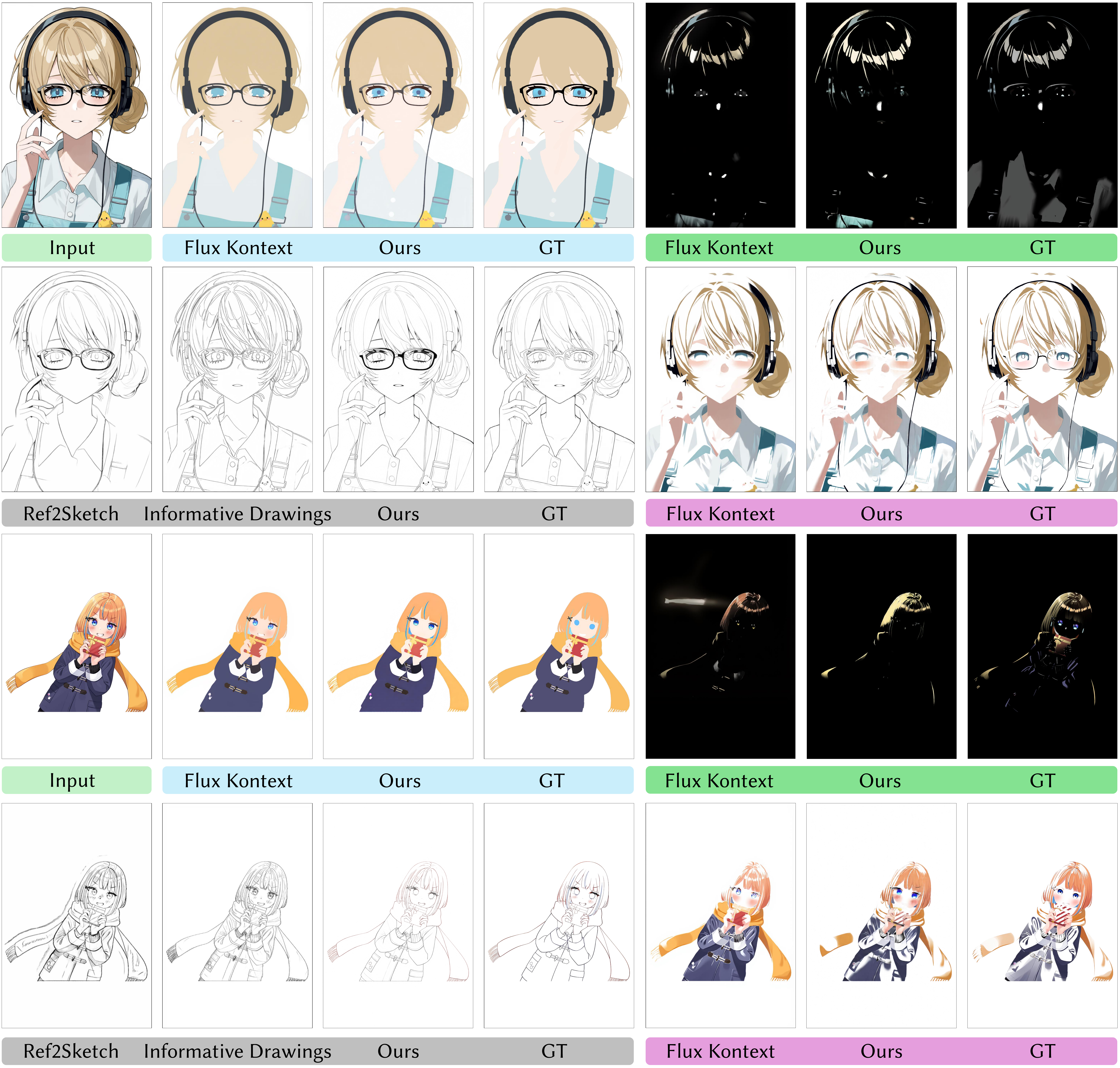}
    \caption{Generalization of our method. For professionally produced animation \bg{inputgreen}{input} with finely layered structures, we compare the \bg{linegray}{line art} results with Ref2Sketch~\cite{seo2023semi} and Informative Drawings~\cite{chan2022learning}, and the \bg{flatblue}{flat color}, \bg{lightgreen}{highlight}, and \bg{renderpurple}{shadow} results with FLUX Kontext~\cite{batifol2025flux}. Copyright: \textcopyright~Live2D Inc.}
    \label{fig:live2d}
\end{figure*}

We provide additional qualitative results to demonstrate the generalization and robustness of our proposed method. Fig.~\ref{fig:live2d} presents two sets of character designs used in actual production, which provide high-quality layering that serves as the GT. As illustrated, our method consistently generates high-quality results that align with the GT across the overall composition. Minor discrepancies remain in certain localized regions, such as the eyes, which are characterized by high complexity and small scale. These deviations are attributed to the fact that our training objective did not specifically target these intricate sub-structures.

\begin{figure*}[!t]
    \centering
    \includegraphics[width=\linewidth]{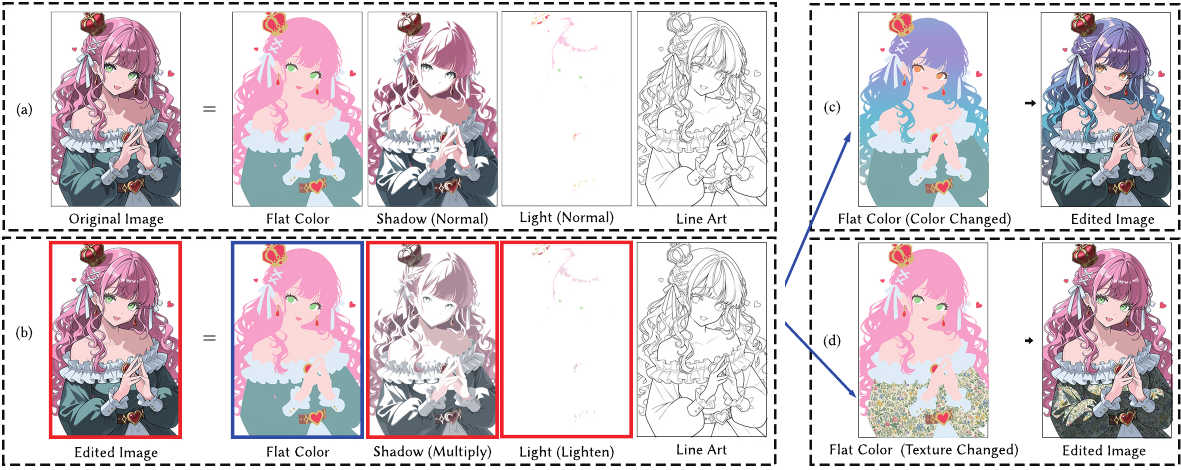}
    \caption{Example of downstream tasks using our method. (a) An illustration can be decomposed into four layers: flat color, shadow, light, and line art. (b) Setting shadow and light layers to Multiply and Lighten modes accurately restores the original rendering. The decomposed layers enable flexible artistic modifications, such as (c) changing flat colors and (d) embedding complex textures.}
    \label{fig:mul_lig_mode}
\end{figure*}

To demonstrate the utility of our framework, we show some applications that significantly streamline professional animation and illustration workflows. Our framework decomposes an anime illustration into distinct layers for line art, flat color, highlight, and shadow. As shown in Fig. \ref{fig:mul_lig_mode}(b), artists can achieve rendering results nearly identical to the original artwork by reassigning the shadow and light layers to standard mathematical blend modes, such as multiply and lighten. The decomposed layers also allow for seamless modifications of the base color while maintaining consistent lighting and shading effects, as illustrated in Fig. \ref{fig:mul_lig_mode}(c). Furthermore, as demonstrated in Fig. \ref{fig:mul_lig_mode}(d), isolating the base color enables creators to directly integrate complex patterns or textures onto a character's clothing without the need for manual warping to match existing shadows, thereby preserving the integrity of the surrounding layers.

Disentangling shading, lighting, and line art layers also opens up potential applications in 2.5D animation and stylized editing, enabling more realistic reposing with proper shadow deformation while allowing independent manipulation of line thickness and style without affecting underlying colors.

\section{Conclusion}

In this work, we presented a novel workflow-aware layer decomposition method for anime illustrations that separates images into line art, flat color, light, and shadow layers, enabling better asset reuse and supporting anime illustration production workflows. To reduce coupling between layers, we introduced a layer semantic embedding and a set of layer-wise losses to guide the decomposition process. We also collected a paired dataset of high-quality character images with layer annotations. Experimental results demonstrated that our method can reliably decompose anime illustrations into meaningful layers. We believe this novel layer decomposition method and the constructed dataset can facilitate future research in anime editing and compositing tasks with better controllability.

\section*{Ackonwledgements}
This work was supported by JST BOOST Program Japan Grant Number JPMJBY24D6, and the research fund from Live2D Inc. corporation. We thank the artists and researchers at Live2D Inc. for providing art design and helpful discussions.

{\small
\bibliographystyle{ieee_fullname}
\bibliography{main}
}

\clearpage
\appendix
\section*{Supplementary Material}

\section{Anime Illustration Layers}

In professional anime production workflows, the decomposition of illustrations into functional layers is essential for a non-destructive and collaborative creative process. The line art layer serves as the geometric foundation by defining the structural topology and semantic boundaries that guide all subsequent rendering stages. Building upon this base, the flat color layer encapsulates the local albedo of each region to act as a pixel-level mask, which allows for precise color adjustments without affecting neighboring areas. To simulate volume and environmental lighting, the shadow layer is isolated to capture ambient occlusion and directional shading for providing depth to the flat regions. Finally, the highlight layer is used to define the properties of the material and specular reflections that include the metallic luster of the accessories or the expressive glints in the eyes of a character.

\begin{figure*}[t]
    \centering
    \includegraphics[width=0.9\linewidth]{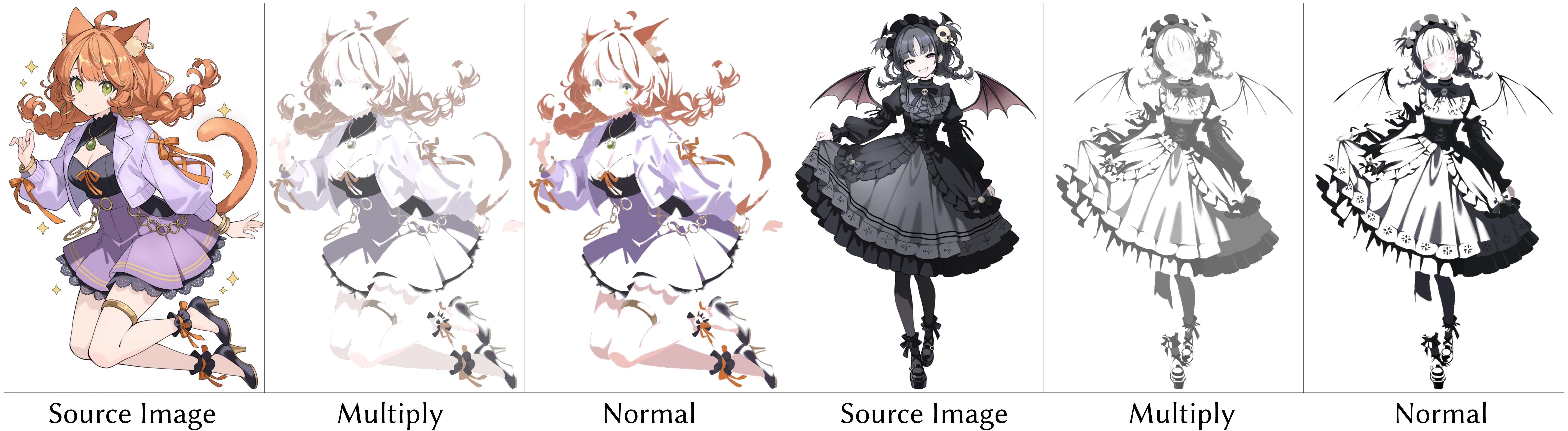}
    \caption{Shadow expressions under different composition schemes.}
    \label{fig:shadow}
\end{figure*}

Specifically, as shown in Fig.~\ref{fig:shadow}, shadow layers exhibit different visual expressions depending on the composition method. In anime illustration, Normal and Multiply are the two most common modes. Normal blending is the most universal and intuitive approach, as it performs direct mixing based on Alpha values. However, if a texture is inserted under a Normal shadow layer, the texture will be obscured by the opaque regions of the shadow. By switching to Multiply mode, the underlying texture can be clearly revealed. Therefore, adopting Multiply blending significantly enhances the versatility of the shadow layer. While these different expressions can be converted into one another, our current method primarily focuses on shadows under Normal composition. We plan to expand our dataset in the future to discuss a wider variety of expression modes.

\section{More Visual Results}

In this section, we present extended dataset samples, comparative studies against state-of-the-art baselines, and more ablation experiments. These results further demonstrate our model’s robustness in extracting functional layers.

\begin{figure}[t]
    \centering
    \includegraphics[width=0.95\linewidth]{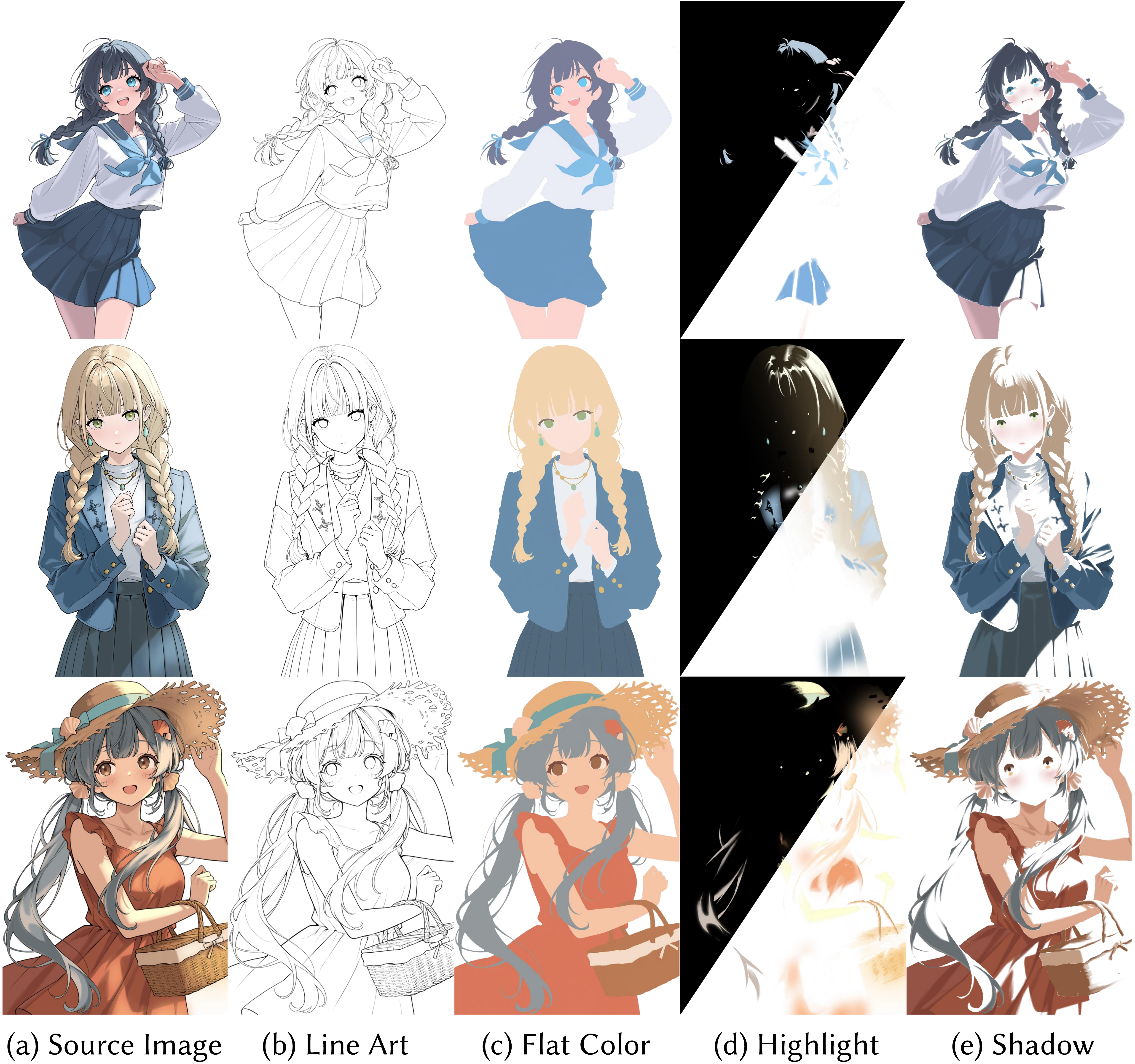}
    \caption{Additional examples of our dataset. To enhance visibility, we show the highlight with a split-black background.}
    \label{fig:more_dataset}
\end{figure}

\textbf{Dataset.} As shown in Fig.~\ref{fig:more_dataset}, we present additional data samples from our dataset. All assets have been meticulously hand-drawn by professional artists and properly organized into four functional layers. The source files are provided in PSD format, which technically allows for high-resolution training. However, to maintain an optimal balance between computational efficiency and output quality, we opted for a training resolution of $768 \times 1152$.

\begin{figure*}
    \centering
    \includegraphics[width=\linewidth]{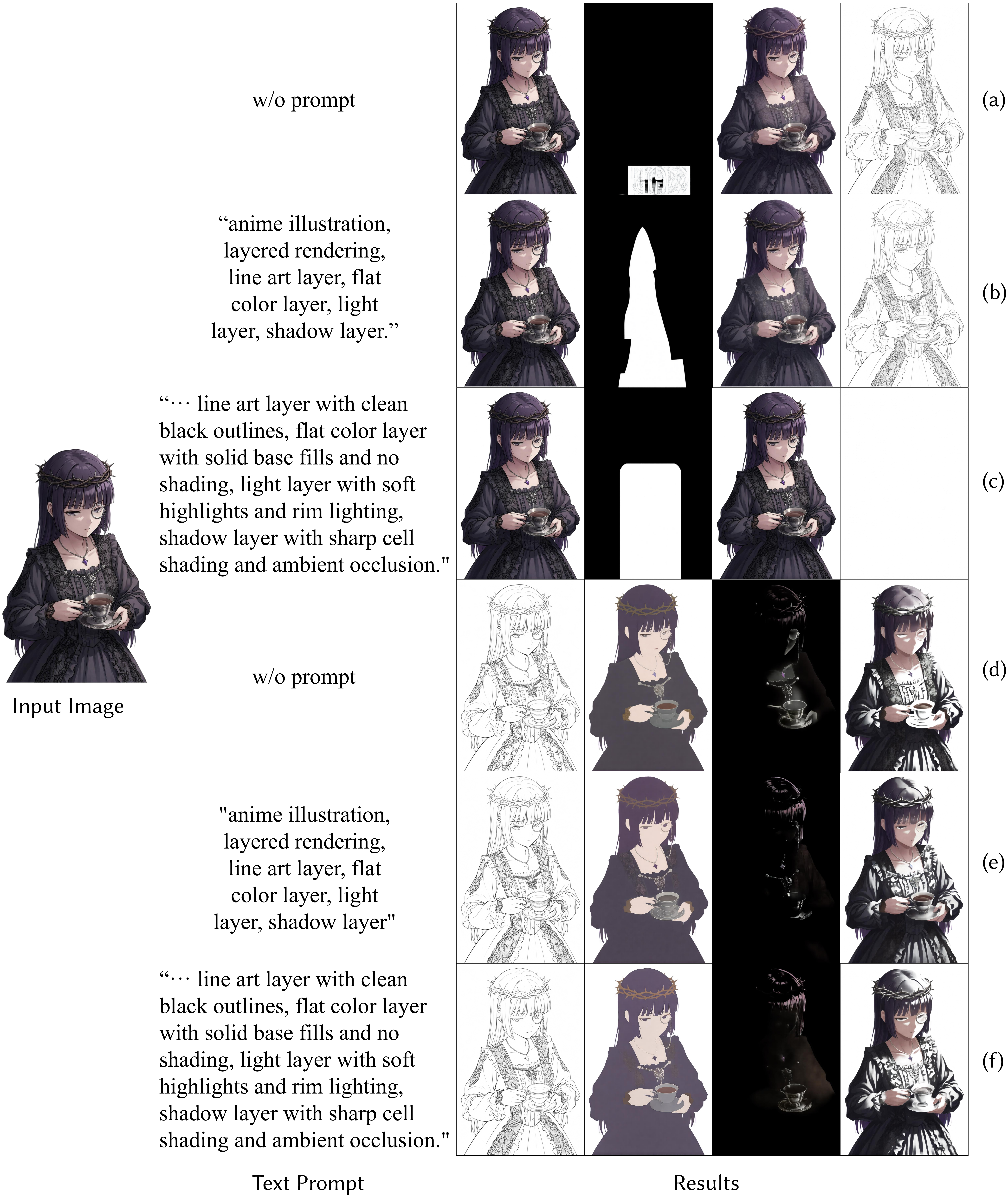}
    \caption{Comparison with the Qwen-Image-Layered~\cite{yin2025qwen} base model. (a)-(c) are generated using the base model with various text prompts, while (d) - (f) shows our results with various text prompts. For images with lighter color palettes, a black background is overlaid to enhance visibility.}
    \label{fig:qil}
\end{figure*}

\textbf{Comparative study.} Fig.~\ref{fig:qil} illustrates a comparison between our approach and the base model, Qwen-Image-Layered~\cite{yin2025qwen}. Specifically, three different text prompt cases are presented for comparison:
\begin{itemize}
    \item Fig.~\ref{fig:qil} (a) and (d) without text prompt.
    \item In Fig.~\ref{fig:qil} (b) and (e), the prompt are set as ``\textit{anime illustration, layered rendering, lineart layer, flat color layer, light layer, shadow layer}''.
    \item In Fig.~\ref{fig:qil} (c) and (f), the prompt are set as ``\textit{high quality anime illustration, masterpiece, layered rendering, lineart layer with clean black outlines, flat color layer with solid base fills and no shading, light layer with soft highlights and rim lighting, shadow layer with sharp cell shading and ambient occlusion}''.
\end{itemize}
As discussed in our paper, models like Qwen-Image-Layered are primarily built upon image segmentation for layer decomposition. These models tend to treat layers as homogeneous components—meaning each layer serves the same functional role as simply a spatial segment of the original image. In contrast, the decomposition of complex anime illustrations involves layers with distinct and heterogeneous functions, significantly increasing the complexity of the task. As demonstrated in Fig.~\ref{fig:qil} (a) - (c), we incrementally increased the complexity of the text prompts. However, in all cases, the baseline models struggled to achieve functional layer decomposition and produced disordered layer sequences that did not align with the input text. Conversely, our method clearly fixes the functional identity of each layer through the layer semantic embedding. Our method also successfully extracts the required functional layers and provides robust support for various downstream applications (as shown in Fig.~\ref{fig:qil} (e)).

\begin{figure}[t]
    \centering
    \includegraphics[width=\linewidth]{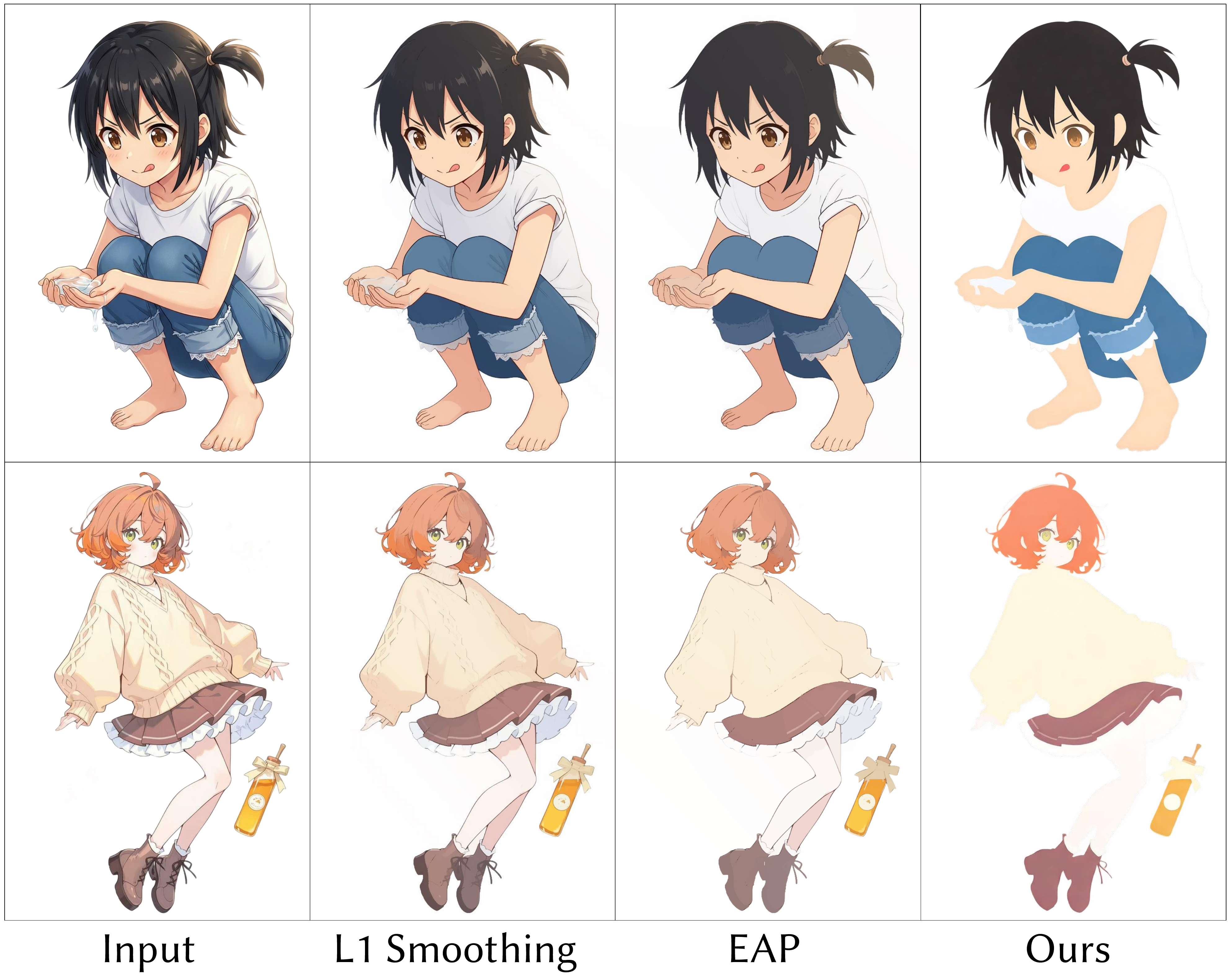}
    \caption{Comparison of flat color outputs with L1 Smoothing~\cite{bi20151} and EAP~\cite{zhang2020erasing}.}
    \label{fig:l1}
\end{figure}

\begin{figure}[t]
    \centering
    \includegraphics[width=\linewidth]{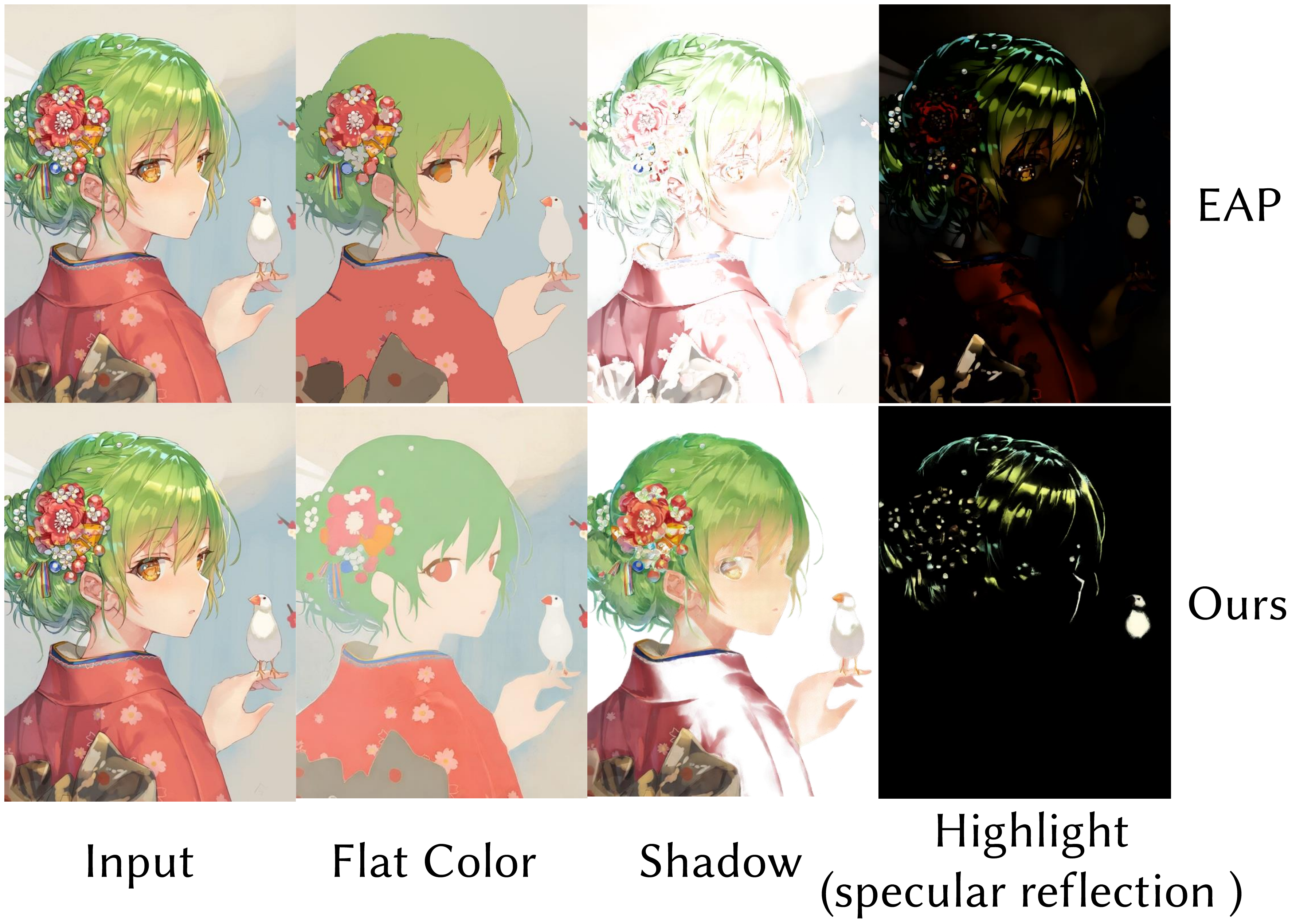}
    \caption{Qualitative comparison of decomposed layers (Flat Color, Shadow, and Highlight) between our method and the representative example of EAP~\cite{zhang2020erasing}.}
    \label{fig:eap}
\end{figure}

We further conducted comparative evaluations against L1 smoothing~\cite{bi20151} and EAP~\cite{zhang2020erasing}. Since both frameworks perform intrinsic decomposition based on L1 smoothing and the source code for this process remains unavailable, we provide a qualitative comparison of the accessible flat color outputs in Fig.~\ref{fig:l1}. As illustrated, both L1 smoothing and EAP tend to retain residual lighting and shading effects during the extraction process. A prominent example is the hair region in the first row. In contrast, our method generates a clean and complete flat color layer. To further distinguish our lighting decomposition from prior work, we performed a layer-wise comparison of the flat color, shadow, and highlight layers using a reference case from the official EAP project page\footnote{\url{https://lllyasviel.github.io/AppearanceEraser/}}. As shown in Fig.~\ref{fig:eap}, EAP fails to properly eliminate the floral ornaments in the flat color layer and suffers from partial background color loss. Discrepancies in the shadow layer primarily stem from differing shading paradigms between the two methods. It should be noted that the reference image has a resolution of $512 \times 700$. Our model produces an output of $512 \times 688$ due to internal architectural constraints.

\textbf{Prompt sensitivity.} As illustrated in Fig.~\ref{fig:qil} (d) - (f), we evaluate the influence of text prompts on our model. As the complexity of the prompts increases, the generated results exhibit only minor variations in lighting and brightness. In nearly all cases, the model consistently generates a complete set of the four functional layers, demonstrating that text prompts have a minimal impact on model stability. Consequently, we recommend using the prompt configuration shown in Fig.~\ref{fig:qil} (e) during inference to maintain consistency with the training.



\section{Possible Applications}

\begin{figure*}[t]
    \centering
    \includegraphics[width=0.94\linewidth]{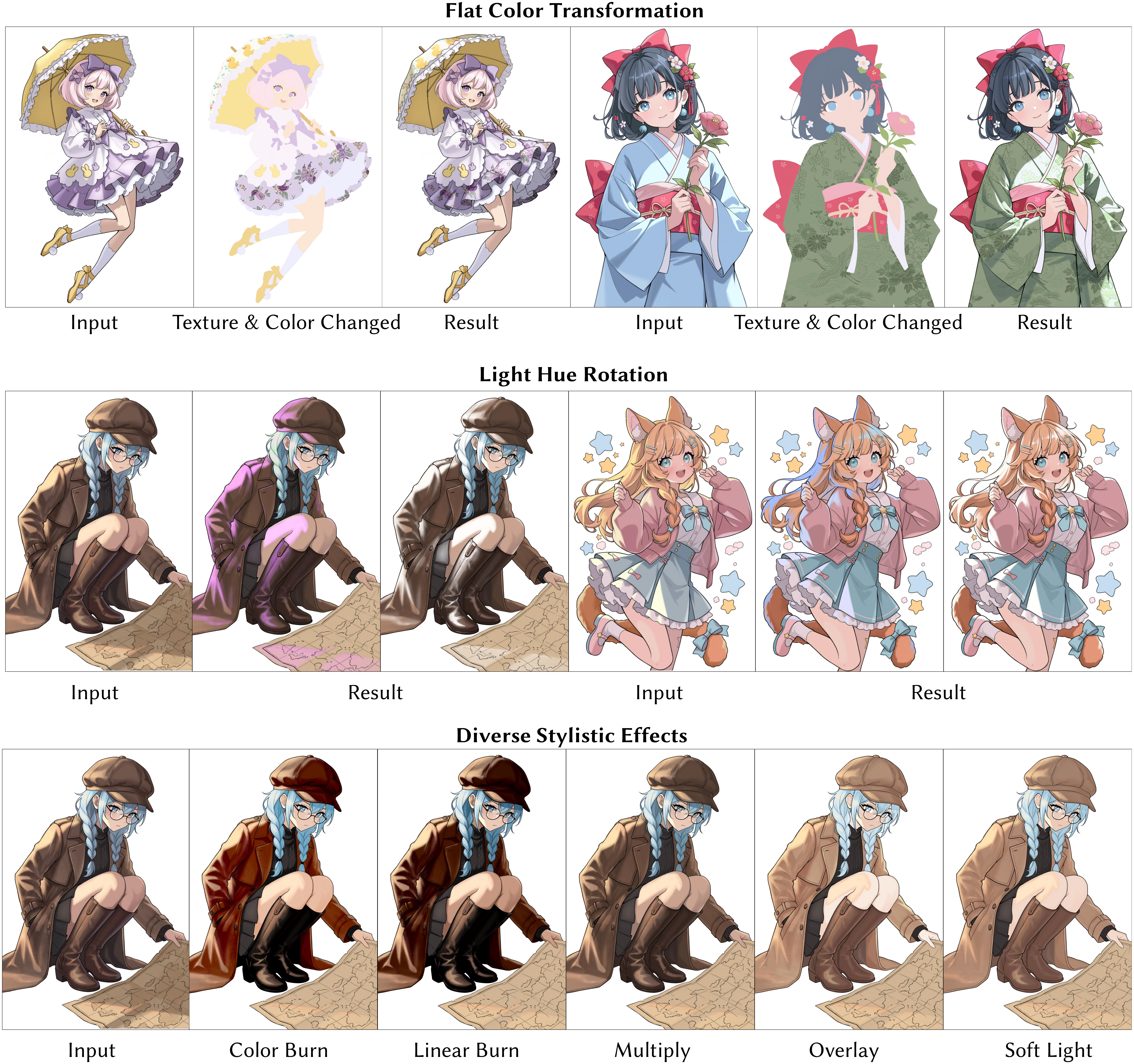}
    \caption{Our decomposed layers allow for precise control over surface attributes like light hue and material textures.}
    \label{fig:app}
\end{figure*}

Our decomposition framework facilitates efficient asset reuse in anime illustration, significantly reducing the manual labor required by artists. Beyond its primary function, our method supports diverse downstream applications. As illustrated in Fig.~\ref{fig:app}, by manipulating the flat color and texture layers, we achieve seamless material and color variations while preserving original shading consistency. Furthermore, our method enables independent hue rotation of the light layers, allowing for sophisticated environmental lighting adjustments without compromising the overall color balance. Moreover, by decomposing the input image and reconfiguring its composition schemes, our framework facilitates a wide range of stylistic variations.

Beyond the demonstrated use cases, our decomposition framework could potentially serve as a cornerstone for several broader research directions in digital illustration. For instance, in the field of automated colorization, the isolated line art and flat color layers might provide explicit structural and semantic priors, hypothetically assisting generative models in achieving more precise color boundary control and reducing common artifacts like color bleeding. Furthermore, the ability to extract clean, functional layers suggests a promising path for sketch simplification and refinement tasks. By treating the line art as a decoupled entity, future researchers could explore more sophisticated stroke-level optimization without the interference of shading or texture.

\end{document}